\documentclass[times,twocolumn,final]{elsarticle}
\usepackage{medima}
\usepackage{framed,multirow}

\usepackage{latexsym}

\usepackage{url}
\usepackage{booktabs}
\usepackage{amssymb, amsmath,amsfonts}

\usepackage{caption}
\usepackage{subcaption}

\usepackage{hyperref}

\definecolor{newcolor}{rgb}{.8,.349,.1}

\journal{Preprint}

\begin{document}

\verso{Kristoffer Wickstrøm \textit{et~al.}}

\begin{frontmatter}

\title{A clinically motivated self-supervised approach for content-based image retrieval of CT liver images}%

\author[1]{Kristoffer Knutsen \snm{Wickstrøm}\corref{cor1}}
\cortext[cor1]{Corresponding author: kristoffer.k.wickstrom@uit.no;}
\author[1]{Eirik Agnalt \snm{Østmo}}
\author[2]{Keyur \snm{Radiya}}
\author[1,2]{Karl Øyvind \snm{Mikalsen}}
\author[1,3]{Michael Christian \snm{Kampffmeyer}}
\author[1,3,4]{Robert \snm{Jenssen}}

\address[1]{Machine Learning Group at the Department of Physics and Technology, UiT the Arctic University of Norway, Tromsø NO-9037, Norway}
\address[2]{Department of Gastrointestinal Surgery, University Hospital of North Norway (UNN), Tromsø, Norway}
\address[3]{Norwegian Computing Center, Department SAMBA, P.O. Box 114 Blindern,
Oslo NO-0314, Norway}
\address[4]{Department of Computer Science, University of Copenhagen, Universitetsparken 1, 2100 København Ø, Denmark}

\begin{abstract}
%%%
Deep learning-based approaches for content-based image retrieval (CBIR) of CT liver images is an active field of research, but suffers from some critical limitations. First, they are heavily reliant on labeled data, which can be challenging and costly to acquire. Second, they lack transparency and explainability, which limits the trustworthiness of deep CBIR systems. We address these limitations by (1) proposing a self-supervised learning framework that incorporates domain-knowledge into the training procedure and (2) providing the first representation learning explainability analysis in the context of CBIR of CT liver images. Results demonstrate improved performance compared to the standard self-supervised approach across several metrics, as well as improved generalisation across datasets. Further, we conduct the first representation learning explainability analysis in the context of CBIR, which reveals new insights into the feature extraction process. Lastly, we perform a case study with cross-examination CBIR that demonstrates the usability of our proposed framework. We believe that our proposed framework could play a vital role in creating trustworthy deep CBIR systems that can successfully take advantage of unlabeled data.
%%%%
\end{abstract}

\end{frontmatter}

%\linenumbers

%% main text
\section{Introduction}
\label{sec1}
Content-based image retrieval (CBIR) is a core research area in medical image analysis, with numerous studies across many different image modalities \citep{dermaXAI, 9296383, HAQ2021101847}. CBIR supports clinicians in retrieving relevant images from a large database compared to a query image, which reduces labor-intensive manual search and aids in diagnosis. For instance, a physician might want to investigate how patients in a large database with a similar disease as a new patient, such as liver metastasis, were diagnosed. The information from the previous diagnoses can then be used to determine the proper treatment for the new patient. In analysis of CT image of the liver, CBIR have been an active and important area of medical image analysis for many years \citep{1403458,liverhandcraft2013,9043172}. CBIR has the potential to make labour intensive tasks in the clinical workflow more time efficient, as illustrated in Section \ref{sec:use-case}.

\begin{figure*}[ht]
    \centering
    \includegraphics[width=0.95\textwidth]{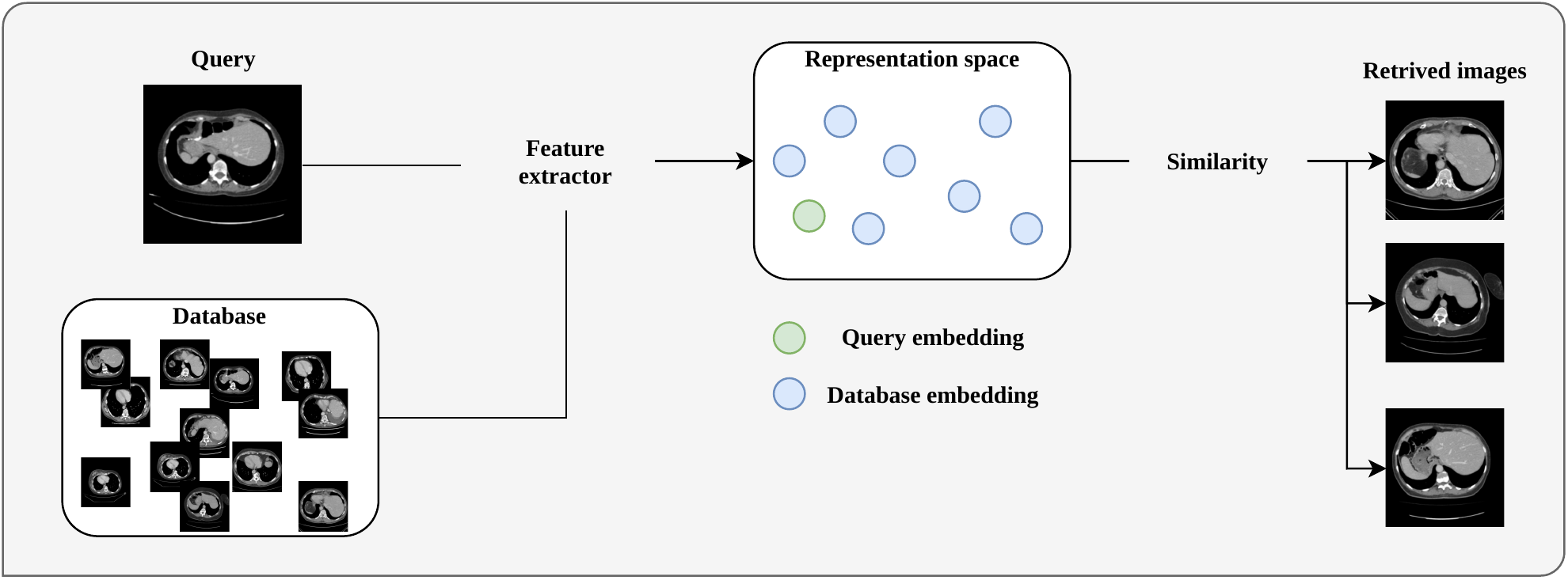}
    \caption{Illustration of content-based image retrieval.}
    \label{fig:cbirc}
\end{figure*}

Currently, deep learning-based CBIR, or deep CBIR, constitute the state-of-the-art of CBIR \citep{guidedCBIR,9043172,HAQ2021101847}, due to its high precision and efficiency. However, deep CBIR suffers from some critical limitations. First (1), current deep CBIR for CT liver images rely on labeled data for training \citep{9043172}. Obtaining labeled data can be costly and time-consuming, which therefore limits the usability of deep CBIR systems. However, recent works have shown how self-supervised learning can leverage unlabeled data to improved CBIR systems \citep{8982468,s22062188}, but such approaches have not been explored in the context of CBIR of CT liver images. Second (2), deep CBIR suffer from a fundamental lack of explainability. This can have detrimental effects in a clinical setting, since deep learning-based systems are known to exploit confounding factors and artifacts to make their predictions. For instance, \cite{gautamisbi} showed that a deep-learning-based system learned to use tokens and artifacts in X-ray images to makes its predictions instead of clinically relevant features. These tokens and artifacts would not be present for new patients, and such a system would not work as intended if put into clinical practice. Therefore, it is not advisable to blindly trust the retrieved images from the deep CBIR system without investigating what input features influence the retrieval process through an explainability analysis.

A promising direction to address the first limitation is learning from unlabeled data through self-supervision. Recent self-supervised learning frameworks have shown remarkable results, in some cases even rivalling supervised learning \citep{simclr, swav, Simsiam}. In a nutshell, contemporary self-supervised approaches train a feature extractor that extract informative representations by exploiting known invariances in the data. These representations can then be used for other tasks, such as CBIR by taking similarities between the new representation to retrieve similar images. These self-supervised approaches have been show to improve performance in the context of chest X-ray \citep{HowTransferableAreSS,bigMedicalSS} and dermatology classification \citep{bigMedicalSS}, organ andcardiac segmentation \citep{hansen2022anomaly}, and whole heart segmentation \citep{dong2021self}, but have yet to be developed for CBIR of CT liver images.

In this paper, we propose a clinically motivated self-supervised framework for CBIR of CT liver images. Our proposed framework incorporates domain knowledge that exploits known properties of the liver, which leads to improved performance compared to well-known self-supervised baselines. Concretely, a novel Houndsfield unit clipping strategy that removes non-liver pixels from the input and allows the system to focus on the liver is incorporated into the self-supervised training. While the focus in this paper is on the liver in CT images, our proposed framework could also be used to focus on other organs by altering how the Houndsfield units are clipped.

For the second limitation, great improvements have been made in the field of explainable artificial intelligence (XAI) over the last couple of years, and numerous studies have shown how XAI can improve the reliability and trustworthiness of deep learning-based systems in healthcare \citep{guidedCBIR, gautamisbi}. However, the majority of these improvements have been in algorithms that can explain models which produce decisions, such as classification or similarity scores. When learning from unlabeled data through e.g. self-supervised learning, such a score or similarity measure might not be available and standard XAI techniques cannot be applied. But the recent field of representation learning explainability \citep{wickstrom2021relax} aims at explaining vector representations, and can therefore tackle the lack of explainability in deep CBIR. But such a representation learning explainability analysis has not been performed in the context of CBIR of CT liver images.

Our contributions are:

\begin{itemize}
    \item A clinically motivated self-supervised framework specifically designed to extract liver specific features.
    \item A novel explainability analysis that explains the representations produced in the feature extraction process.
    \item Thorough evaluation on real-world datasets.
    \item A case-study where images from the same patient are retrieved across different examinations.
\end{itemize}

\section{Related work}

\subsection{Content-based image retrieval}

The goal of content-based image retrieval (CBIR) is to find similar images from a large-scale database, given a query image. CBIR is an active area of research that span numerous medical imaging domains, such as X-ray \citep{HAQ2021101847, guidedCBIR}, dermatology \citep{dermaXAI, Ballerini2010}, mammography \citep{Jiang2014}, and histopathology \citep{Peng2019, Zheng2019}. An illustration of a CBIR system in the context of CT liver images is shown in Figure \ref{fig:cbirc}.

\subsection{Content-based image retrieval of CT liver images}

CBIR of CT liver images have been extensively studied. Early studies relied on handcrafting features based on certain properties in the images. Gabor filters have been used to extract texture information \citep{1403458}. Texture information have also been combined with density information in the context of focal liver lesion retrieval \citep{liverhandcraft2013}. Histogram-based features extraction have been explored to retrieve CT scans with similar liver lesions. Manifold learning have been utilized to facilitate CBIR of CT liver images \citep{Mirasadi2019}. Lastly, a Bayesian approach has been studied in connection with multi-labeled CBIR of CT liver images \citep{9296383}.

Recently, deep learning-based feature extraction have improved performance significantly in CBIR of CT liver images. The most straight forward approach for deep CBIR is to train a neural network for the task of CT liver image classification and use the intermediate features prior to the classification layer for calculating similarities. This has been demonstrated to produce good results when the network was trained for the task of focal liver lesions detection \citep{9043172}. However, all these approaches need labeled data to train the deep learning-based feature extractor.

\subsection{Self-supervised learning}

Learning from unlabeled data is a fundamental problem in machine learning. Recently, self-supervised learning have shown promising results in computer vision \citep{simclr, Simsiam}, natural language processing \citep{BERT, GPT}, and time series analysis \citep{franceschi, WICKSTROM202254}. Furthermore, recent studies have also demonstrated that self-supervised learning can improve performance across several imaging domains in medical image analysis \citep{bigMedicalSS, HowTransferableAreSS,hansen2022anomaly,dong2021self}.

For computer vision, there are three main approaches to self-supervised learning. First, contrastive self-supervised learning is performed by sampling positive pairs and negative samples and learning a representation where the positive pairs are mapped in close proximity and far from the negative samples. The SimCLR framework \citep{simclr} is one of the most widely used approaches in this category. Second, clustering-based self-supervised learning utilizes clustering algorithms to produce pseudo-labels which in turn are used to learn a useful representation of the data. DeepCluster \citep{DeepCluster} and the SwAV framework \citep{swav} are two of the most widely used clustering-based self-supervised approaches in the literature. Lastly, siamese self-supervised approaches learns how to produce a useful representation by maximizing agreement between positive pairs of samples. The two main contemporary approaches in siamese self-supervised approaches is the SimSiam framework \citep{Simsiam} and the BYOL framework \citep{byol}.

\subsection{Explainability}

Explainability is of vital importance for machine learning systems in healthcare. Without it, clinicians cannot fully trust the algorithms decision and the system becomes less reliable. Many recent studies have shown how explainability can be incorporated into deep learning systems for medical image analysis, ranging from diabetic retinopathy \citep{DRxai}, dermatology \citep{dermaXAI, 9246575}, X-ray \citep{xrayXAI}, and endoscopic images \citep{WICKSTROM2020101619, endocopyXAI}.

Most of the widely used explainability techniques typically leverage the classification or similarity score to ascertain input feature importance \citep{guidebackprop,Schulz2020Restricting,Plummer2020}, and such approaches have been explored in the context of deep CBIR. For models trained for classification tasks, explanations through gradient information have been shown to both provide new insights and improve performance for X-ray images \citep{guidedCBIR}. For models trained to output a similarity score, it has been shown how the similarity score can be used to provide explanations \citep{CvprDong2019, Plummer2020}. Similarity score explanations have been explored for X-ray images \citep{9706900}. Lastly, it has been shown that explanation by examples can be effective in histopathological images \citep{Peng2019}.

In the unlabeled setting where only the feature extraction model is available, these techniques are not applicable. In such cases, it is desirable to explain the vector representation of an image, since the decision is not available. Representation learning explainability is a very recent field of XAI, that has yet to be developed for CBIR. In this work, we leverage the RELAX framework \citep{wickstrom2021relax} to explain the feature extractors trained using self-supervised learning. RELAX is the first method that allows for representation learning explainability and has been shown to provide superior performance to competing alternatives \citep{wickstrom2021relax}.

\section{A clinically motivated self-supervised approach for CT liver images}

In this section, we present our proposed clinically-motivated self-supervised approach and the SimSiam framework for self-supervised learning.

\subsection{A clinically motivated self-supervised approach for CT liver images}
We propose to incorporate clinical knowledge into our self-supervised framework to learn more clinically relevant features. In self-supervised learning, known invariances in the data are used to train a feature extractor that extracts relevant features from the input images. For instance, the liver can occur on both the left and right hand side of an image, depending on which direction the patient is inspected. Therefore, the feature extractor should be invariant to horizontal flips in the images, and this invariance can be learned by incorporating horizontal flipping into the self-supervised learning procedure. Identifying these invariances is crucial to make the self-supervised system work properly and focus on clinically relevant features in the input images. Our motivation is based on the knowledge that the pixel intensities of the liver lay within a certain range for CT images. A standard pre-processing step is to clip the pixel intensities of the CT images \citep{Li2018}, such that unimportant pixels are removed prior to learning. The pixel intensities of CT images represent a physical quantity, namely the Houndfield unit. The same clipping is usually applied to all images. However, if this clipping was incorporated into the self-supervised learning procedure, the network could be guided to learn which feature are liver features and which ones are not. In a sense, we are exploiting the knowledge that the liver should be invariant to pixel intensity clipping for a certain range of clipping.

\begin{figure}[htb]
    \centering
    \includegraphics[width=0.975\columnwidth]{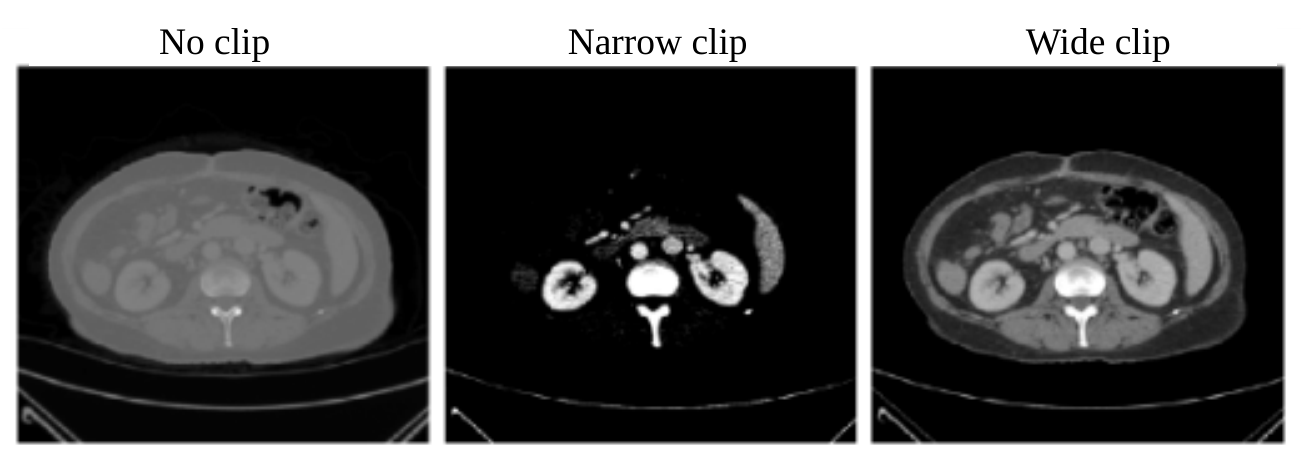}
    \caption{Effect of Houndsfield unit clipping on CT liver images. From left to right, no clipping, narrow clip (50, 150), and wide clip (-200, 300).}
    \label{fig:clip}
\end{figure}

\begin{figure}[ht]
    \centering
    \includegraphics[width=0.95\columnwidth]{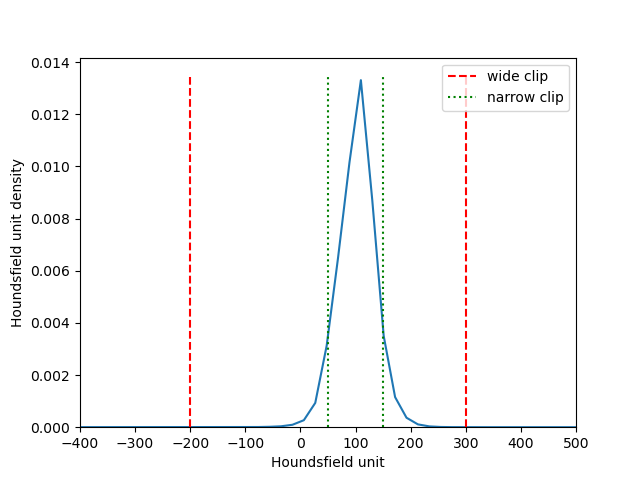}
    \caption{Distribution of pixel intensity values for liver pixel from the Decathlon dataset and the two clipping strategies used in our proposed framework.}
    \label{fig:pixel-i}
\end{figure}

\begin{figure*}[ht]
    \centering
    \includegraphics[width=0.95\textwidth]{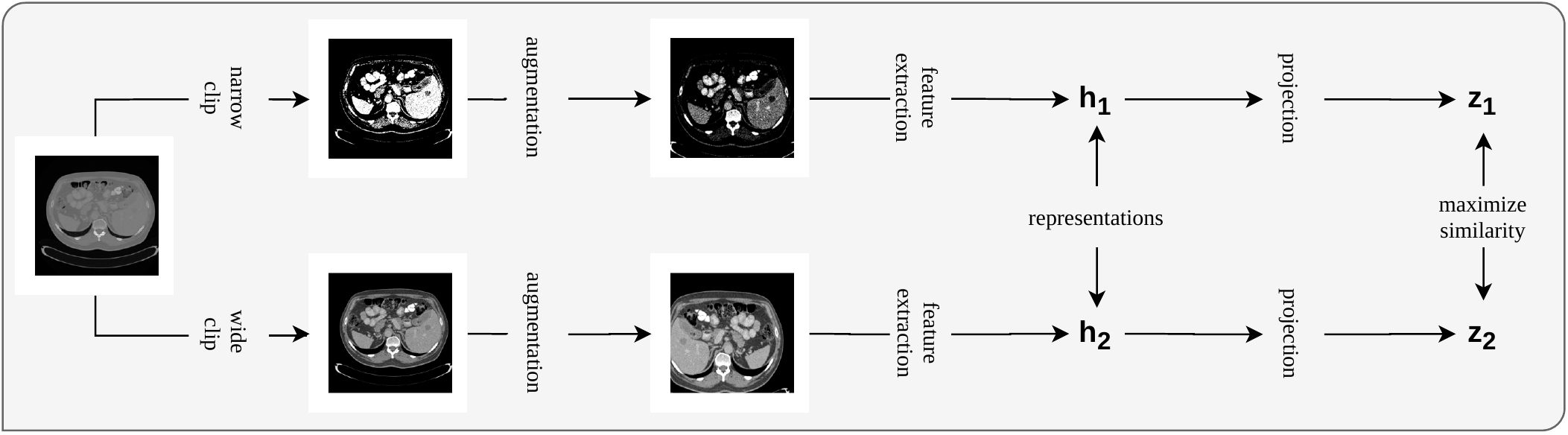}
    \caption{Illustration of proposed self-supervised framework.}
    \label{fig:fwork}
\end{figure*}

Based on this motivation, we propose a Houndsfield clipping strategy where the pixel values for the same image are clipped and scaled based on different ranges of Houndsfield units. Figure \ref{fig:clip} shows how our proposed clipping scheme affects an image. The leftmost image has no clipping applied, and illustrates why it is important to remove some pixel intensities in order to highlight relevant structures in the images. The middle images show the narrow clipping strategy between 50 and 150 Houndsfield units. Notice how only the liver and some other organs are now visible in the image. The rightmost image shows the wide clipping strategy between -200 and 300 Houndsfield units. In this case, some redundant structures are removed, but more organs are left visible compared to the middle image. The images considered in this paper are intra venous contrast enhanced images taken in the portal venous phase. These two ranges were chosen based on the following. First, it is known that the liver typically has Houndsfield units in the range 50-60 \citep{Tisch2019}. Furthermore, we have collected all pixel intensities for the liver in the Decathlon dataset. These values are shown in Figure \ref{fig:pixel-i}, and illustrates how the narrow clip will remove some of the liver pixels but keep the main proportion, while the wide clip will keep almost all liver pixels apart from some outliers. Our proposed framework for learning representations that focus on liver features is shown in Figure \ref{fig:fwork}. Each image is clipped with the wide and narrow range, before the data augmentation is applied. Afterwards, we follow the SimSiam approach described below. During testing, we use the wide clipping to ensure that most liver pixels are kept in the images.

\subsection{SimSiam framework}

In this work, we build on the SimSiam framework. The main motivation for this choice is that both contrastive and clustering-based self-supervised approaches requires a large batch size during training to provide high quality representations \citep{simclr, swav}. This can be computationally challenging, especially if the medical images in question are large. However, the siamese-approaches \citep{Simsiam} are less sensitive to the batch size used during training. Furthermore, we opt for the SimSiam approach over BYOL to avoid training both a student and a teacher network used in BYOL, again to avoid additional computational overhead.

Let $\mathbf{X}\in \mathbb{R}^{H\times W}$ represent an input image with height $H$ and width $W$ and $f$ a feature extractor that transforms $\mathbf{X}$ into a new $d$-dimensional representation $\mathbf{h}\in \mathbb{R}^{d}$, that is $f(\mathbf{X})=\mathbf{h}$. Next, two views $\mathbf{X}_1$ and $\mathbf{X}_2$ are constructed by augmenting the original image. The task performed in SimSiam to learn a useful representation, is to maximize the similarity between the two views. The representation $h$ is the new representation that can be used for downstream tasks, such as the CBIR. However, the loss is not computed directly on the output of the feature extractor $f$. Instead, a multilayer perceptron-based projection head $g$ transforms $\mathbf{h}$ into a new representation $\mathbf{z}$, that is $g(\mathbf{h})=\mathbf{z}$, where the loss is computed. This projector is a crucial component in most self-supervised frameworks \citep{simclr, moco}, as it avoid dimensional collapse in the representation $h$ \citep{jing2022understanding},  which is the one that will be used for downstream tasks such as CBIR. The learning is performed by minimizing the negative cosine similarity between the two views:

\begin{equation}\label{eq:dist}
    D(\mathbf{z}_1, \mathbf{z}_2) = -\frac{\mathbf{z}_1}{\lVert \mathbf{z}_1 \rVert_2}\cdot\frac{\mathbf{h}_2}{\lVert \mathbf{h}_2 \rVert_2},
\end{equation}
where $\lVert \cdot \rVert_2$ denotes the $\ell_2$-norm. 

\begin{equation}\label{eq:loss1}
    L = D(\mathbf{z}_1, \mathbf{h}_2) + D(\mathbf{z}_2, \mathbf{h}_1)
\end{equation}
An important component of the the SimSiam framework is a stop-gradient $\text{{\fontfamily{tnr}\selectfont
(stopgrad)}}$ operation, which is incorporate in Equation \ref{eq:loss1} as follows:

\begin{equation}\label{eq:loss2}
    L = \frac{1}{2}D(\mathbf{z}_1, \text{{\fontfamily{tnr}\selectfont
stopgrad}}(\mathbf{h}_2)) + \frac{1}{2}D(\mathbf{z}_2, \text{{\fontfamily{tnr}\selectfont
stopgrad}}(\mathbf{h}_1))
\end{equation}
The stop-gradient operation is applied to the projector network, such that the encoder on $\mathbf{X}_2$ no gradient from $\mathbf{h}_2$ in the first term, but it recieves gradients from $\mathbf{z}_2$ (and similarly for $\mathbf{X}_1$). The stop-grad operation allows SimSiam to mimic a teacher-student setup, but avoid the need to store two networks. Furthermore, it has been shown that the stop-grad operation is critical to avoid the problem of complete collapse in the representations \citep{collapse}.

\paragraph{Data augmentation} The prior knowledge inject through the data augmentation is of paramount importance to ensure that the models learns relevant features. The data augmentation used in SimSiam is similar to the standard approach in recent self-supervised learning \citep{moco,simclr}:

\begin{enumerate}
    \item Crop with a random proportion from [0.2, 1.0], and resize to a fixed size.
    \item Flip horizontally with a probability of 0.5.
    \item Color augmentation is performed by randomly adjusting the brightness, contrast, saturation, and hue of each image with a strength of [0.4, 0.4, 0.4, 0.1]
    \item Randomly convert image to gray scale version with a probability 0.2.
\end{enumerate}

Note that the input images are converted to pseudo RBG images by stacking the input image 3 times along the channel axis. Prior works have shown that the augmentation scheme listed above can lead to increased performance across several medical image related tasks \citep{bigMedicalSS, HowTransferableAreSS, hansen2022anomaly, dong2021self}, albeit not in the context of CBIR of CT liver images. However, these augmentations are selected with natural images in mind, and do not take into account the properties of CT liver images. Our proposed Houndsfield unit clipping scheme takes into account the particular characteristics of CT images of the liver, which we hypothesize can improve the self-supervised framework.

\section{Explaining representations}

Explainability is a critical component for creating trustworthy and reliable deep learning-based systems. For deep CBIR, we want to know what information the feature extractor is using to create the representation that the retrieval is based on. This requires explaining the vector representations produced by the feature extractor, which can not be accomplished with standard explainability techniques since they require a classification or similarity score to create the explanation. However, the recent field of representation learning explainability address the problem of explaining representations \citep{wickstrom2021relax}. In this work, we leverage the RELAX \citep{wickstrom2021relax} framework to explain the representations used in the CBIR system.

\subsection{RELAX}\label{sec:relax}
\begin{figure}
    \centering
    \includegraphics[width=0.9\columnwidth]{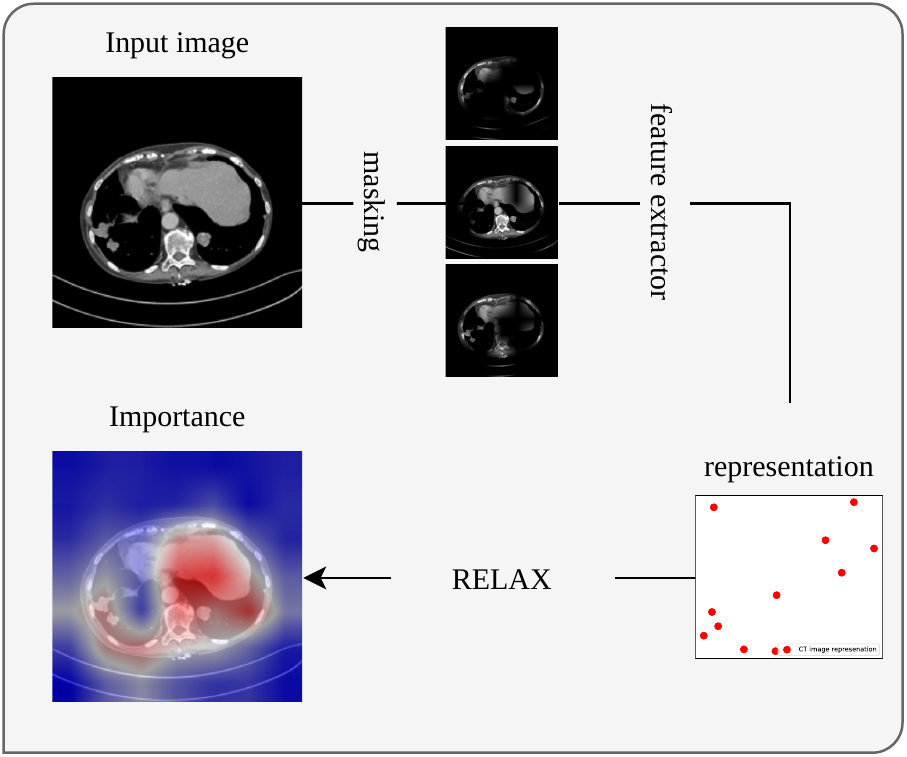}
    \caption{Illustration of RELAX. A feature extractor produces a new representation of an input image, and RELAX determines what input features are important for the representation.}
    \label{fig:relax}
\end{figure}

RELAX is an occlusion-based explainability framework that provides input feature importance in relation to a vector representation, as opposed to a classification or similarity score. The core idea of RELAX is to evaluate how the representation of an image changes as parts of the image are removed using a mask. Let $\mathbf{M}\in [0, 1]^{H\times W}$ represent a stochastic mask used for removing parts of the image. Next, $\bar{\mathbf{h}} = f(\mathbf{X} \odot \mathbf{M})$, where $\odot$ denotes element-wise multiplication, is the representation of a masked version of $\mathbf{X}$ and $s(\mathbf{h}, \bar{\mathbf{h}})$ is a similarity measure between the unmasked and the masked representation. The intuition behind RELAX is that when informative parts are masked out, the similarity between the two representations should be low, and vice versa for non-informative parts. Finally, the importance $R_{ij}$ of pixel $(i,j)$ is defined as:

\begin{equation}\label{eq:relax}
    \bar{R}_{ij} = \frac{1}{N}\sum\limits_{n=1}^N s(\mathbf{h}, \bar{\mathbf{h}}_n)M_{ij}(n).
\end{equation}
Here, $\bar{\mathbf{h}}_n$ is the representation of the image masked with mask $n$, and 
$M_{ij}(n)$ the value of element $(i, j)$ for mask $n$. The similarity measure used in the cosine similarity, as proposed in prior works \cite{wickstrom2021relax}. The RELAX framework is illustrated in Figure \ref{fig:relax}.

The mask generation is a crucial component in RELAX. In this work, we follow the strategy used in previous studies \citep{Petsiuk2018rise, wickstrom2021relax}. Binary masks of size $h\times w$, where $h<H$ and $w<W$, are generated, where each element of the mask is sampled from a Bernoulli distribution with probability $p$. To produce smooth and spatially coherent masks, the small masks are upsampled using bilinear interpolation to the same size as the input image. Furthermore, the number of masks required to obtain reliable estimates of importance is an important hyperparameter. In this work, we generate 3000 masks to obtain an explanation for a single image, as suggested in a prior work \citep{wickstrom2021relax}.

\begin{figure*}[ht]
     \centering
     \begin{subfigure}[b]{0.3\textwidth}
         \centering
         \includegraphics[width=\textwidth]{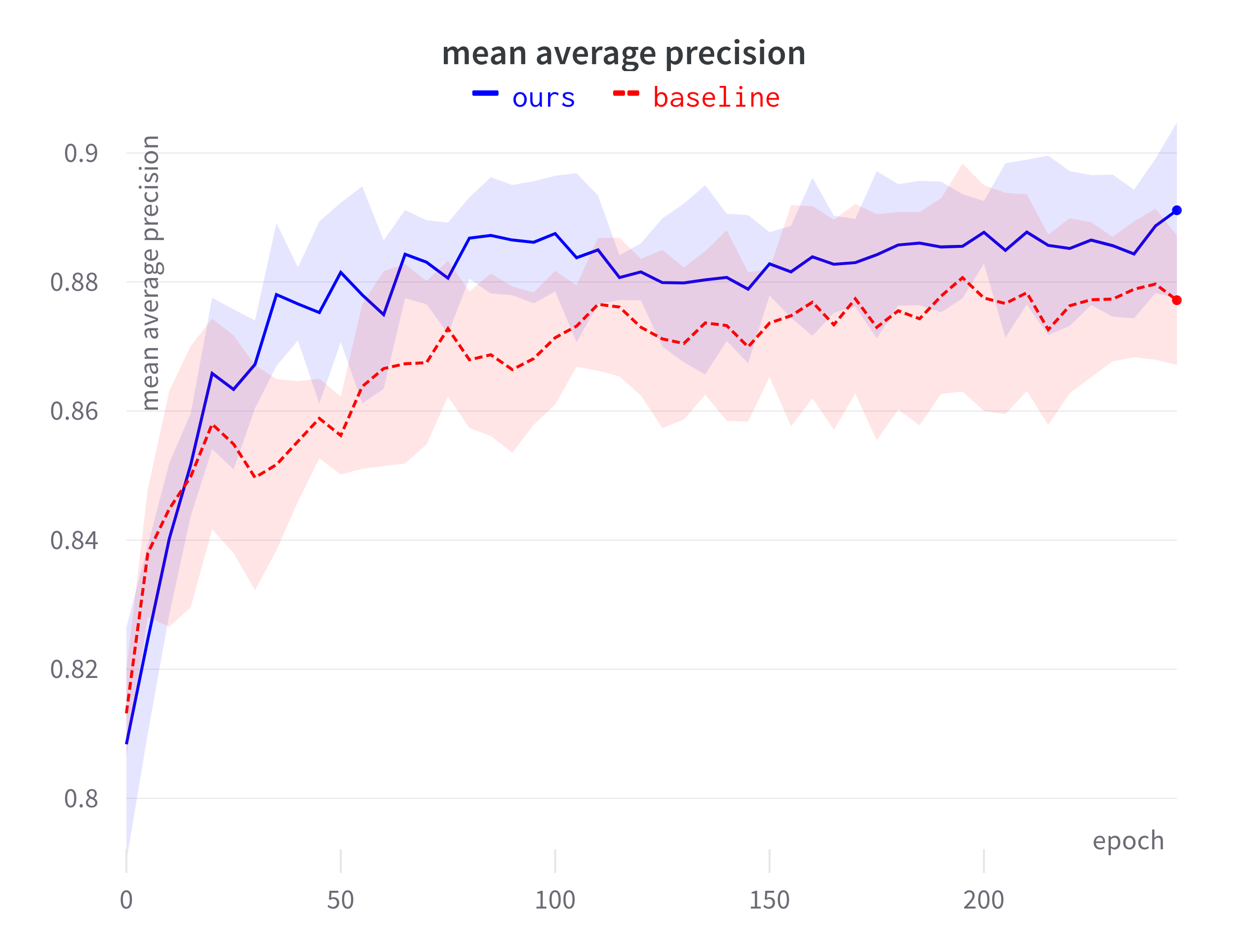}
         \label{fig:unn-map}
     \end{subfigure}
     \hfill
     \begin{subfigure}[b]{0.3\textwidth}
         \centering
         \includegraphics[width=\textwidth]{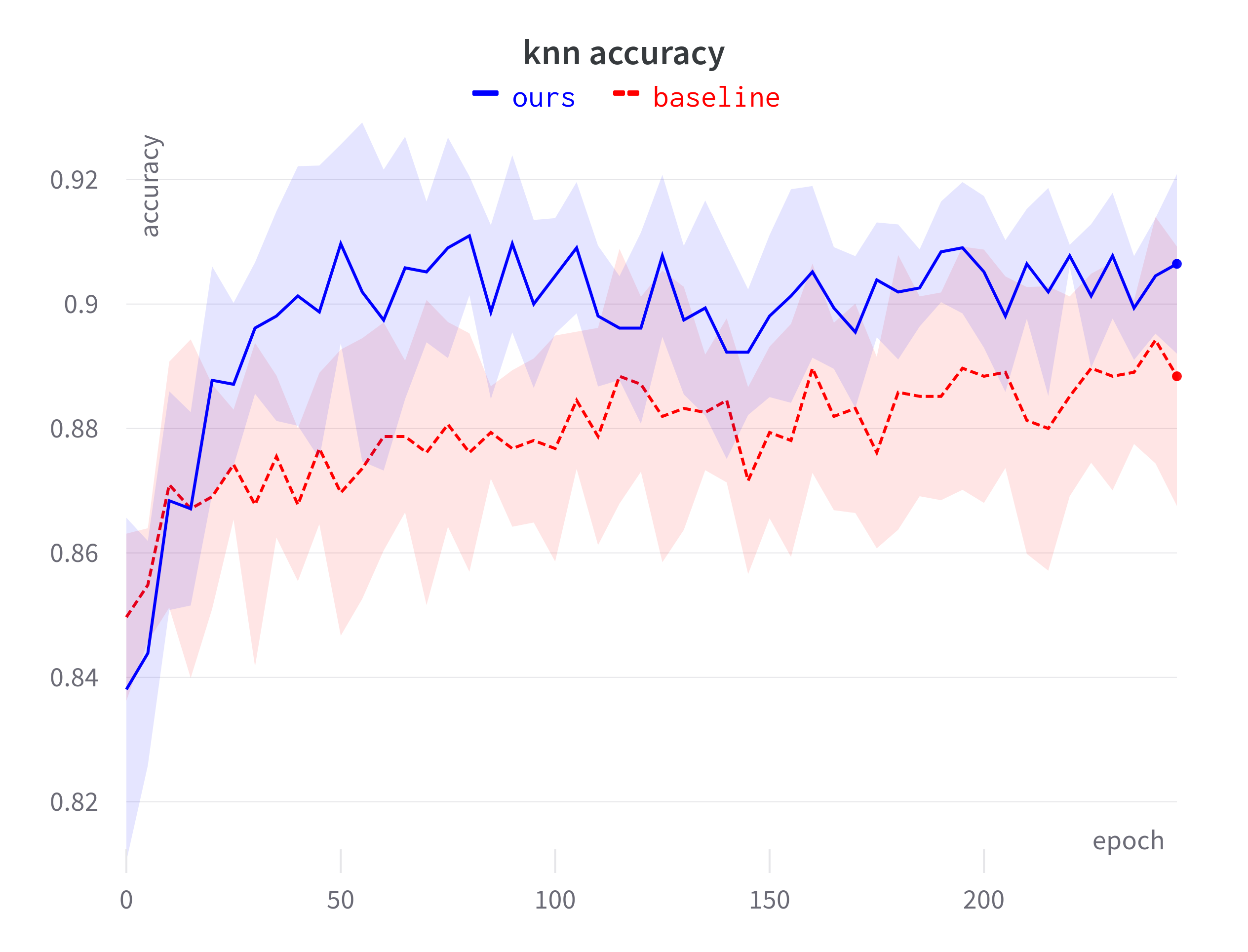}
         \label{fig:unn-acc}
     \end{subfigure}
     \hfill
     \begin{subfigure}[b]{0.3\textwidth}
         \centering
         \includegraphics[width=\textwidth]{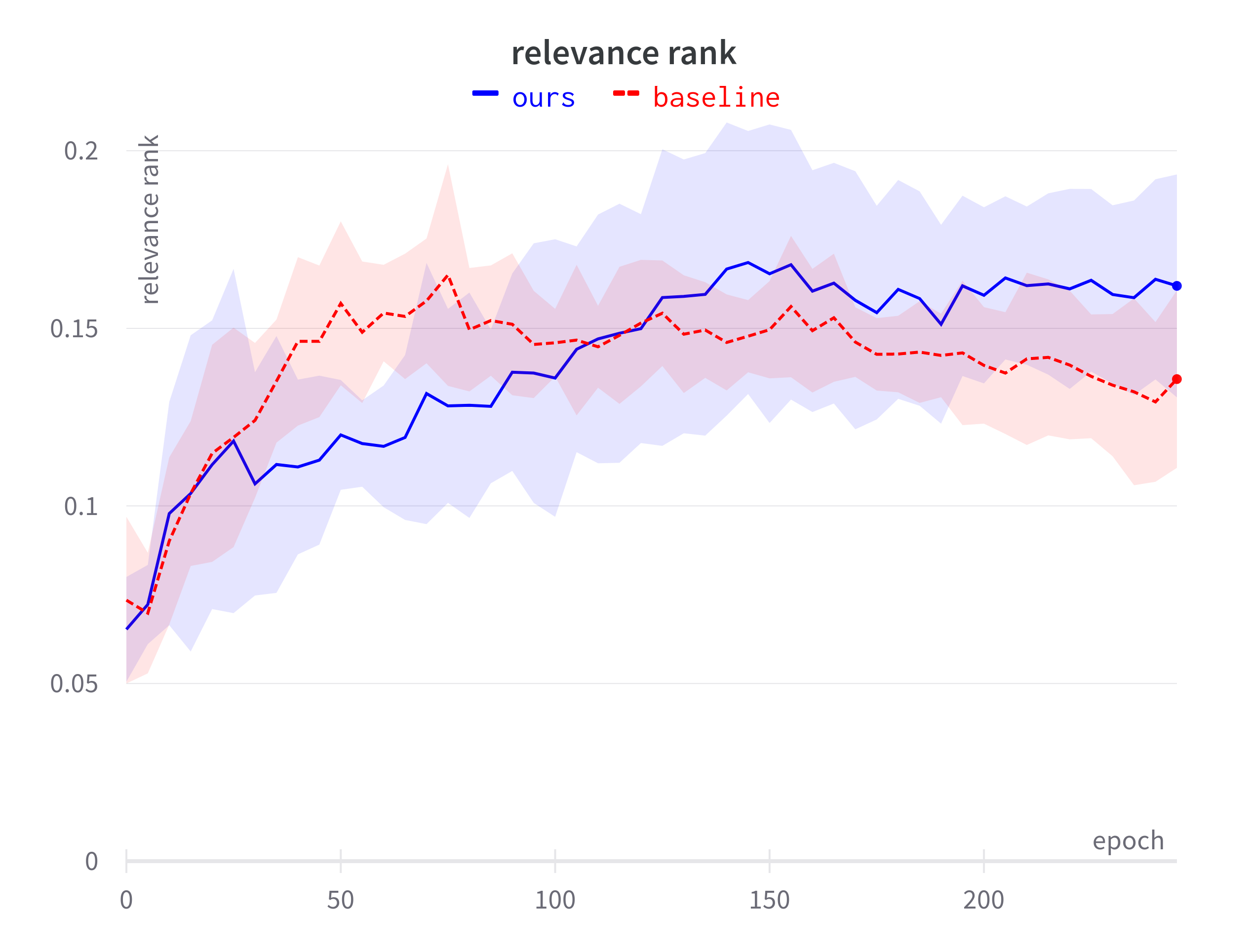}
         \label{fig:unn-rr}
     \end{subfigure}
        \caption{From left to right, mean average precision, knn accuracy, and relevance rank scores versus epochs across 5 training runs on the test images from the Decathlon dataset The plot show how performance increase with training time, and that the proposed framework learns faster with better results.}
        \label{fig:scores}
\end{figure*}

\section{Evaluation}
We introduce the set of scores utilized to provided quantitative evaluation of our proposed framework.

\subsection{Evaluating quality of CBIR}

A standard approach to evaluate the quality of a CBIR system is to measure the class-consistency in the top retrieved images \citep{guidedCBIR, LI201866}. One of the most common approaches to evaluate the class-consistency is through mean average precision (MAP):

\begin{equation}
    \text{MAP} = \frac{1}{N}\sum\limits_{n=1}^N\frac{1}{K}\sum\limits_{k=1}^K\text{precision(k)}_n,
\end{equation}
where $N$ is the number of test samples (query images), $K$ is the top-$K$ retrieved images for each query image, and precision is defined as:

\begin{equation}
    \text{precision(k)} = \frac{\lvert \text{relevant images} \cap \text{k-retrieved images}\rvert}{\lvert \text{k-retrieved images} \rvert}.
\end{equation}
MAP evaluates the precision of the retrieved images across several values of K, which makes it robust towards fluctuations among the top retrieved images.

\subsection{Evaluating quality of representations}

The most widespread approach for evaluating the representation produced by a self-supervised learning framework is to train a simple classifier on the learned representations \citep{simclr,swav,moco}. The motivation for this, is that a simple classifier is highly dependent on the representation it is given in order to perform the desired task. In this work, we follow recent studies that use a k-nearest neighbors (KNN) classifier \citep{caron2021emerging, swav} to evaluate the quality of the representation. We opt for a KNN classifier over a linear classifier as it does not require any training, which can lead to ambiguities in the results \citep{Kolesnikov2019CVPR}, and has minimal hyperparameters to tune.

\subsection{Evaluating the quality of explanations}

Great improvements have been made in the field of XAI over the last couple of years. In contrast, the field of evaluation for explanations is still under active development \citep{doshivelez2017rigorous}. However, recent advances have introduced new methods for providing quantitative evaluation of explanations. In this work, we use the relevance rank accuracy score (RR) \citep{ARRAS202214}. RR measures how many of the top-$M$ relevant pixels lies within the ground truth segmentation mask. It can be considered a proxy for how well the explanation agrees with a human explanation for a given images. Let $R_M$ denote the $M$ most relevant pixels in an explanation, and $S$ the segmentation mask for the liver. RR can then be defined as:

\begin{equation}
    \text{RR} = \frac{1}{N}\sum\limits_n^N \frac{\lvert R_M(n) \cap S(n)  \rvert}{\lvert S(n) \rvert}.
\end{equation}
The RR is computed using the Quantus toolbox \citep{hedstrom2022quantus}.

\begin{figure*}[ht]
     \centering
     \begin{subfigure}[b]{0.3\textwidth}
         \centering
         \includegraphics[width=\textwidth]{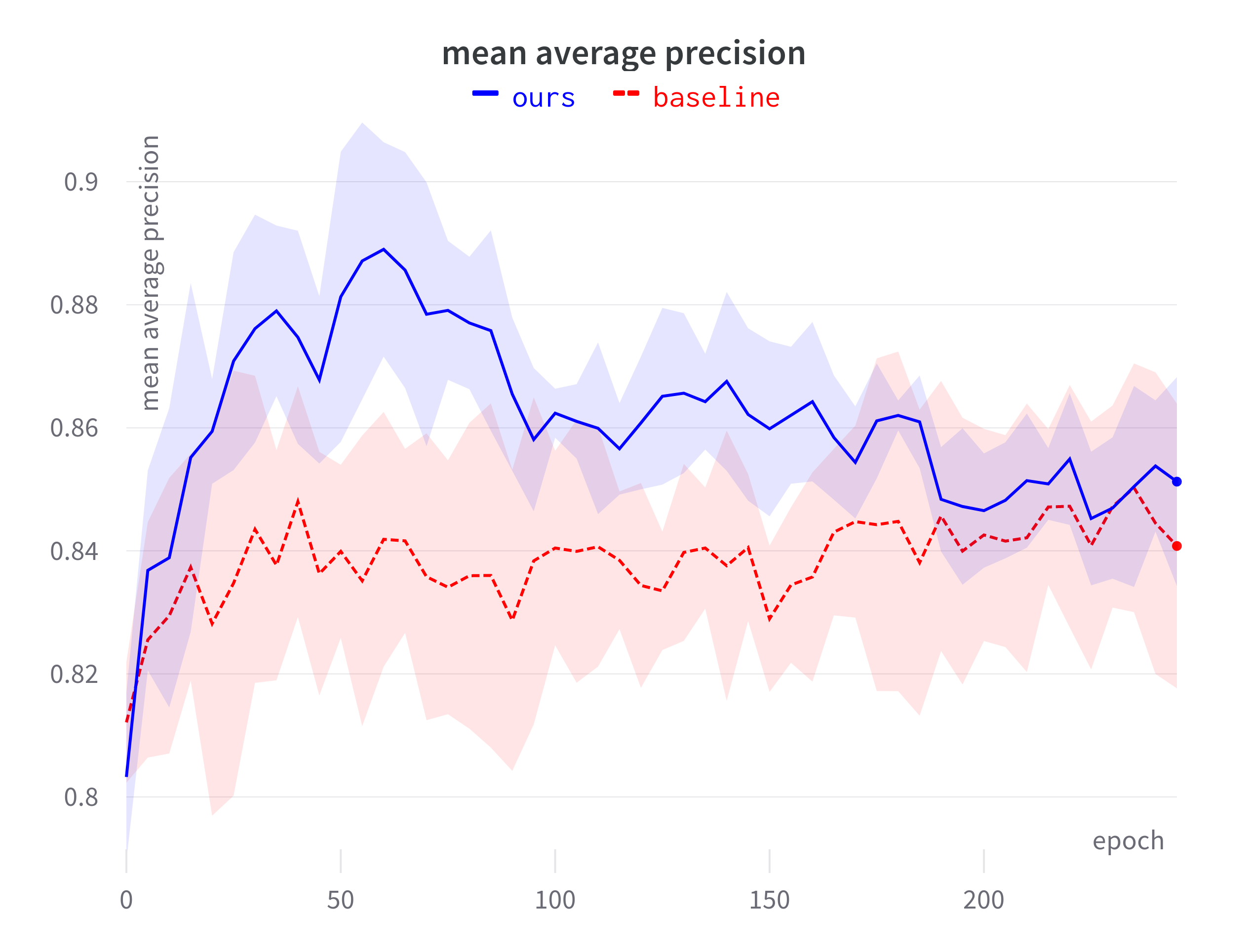}
         \label{fig:map}
     \end{subfigure}
     \hfill
     \begin{subfigure}[b]{0.3\textwidth}
         \centering
         \includegraphics[width=\textwidth]{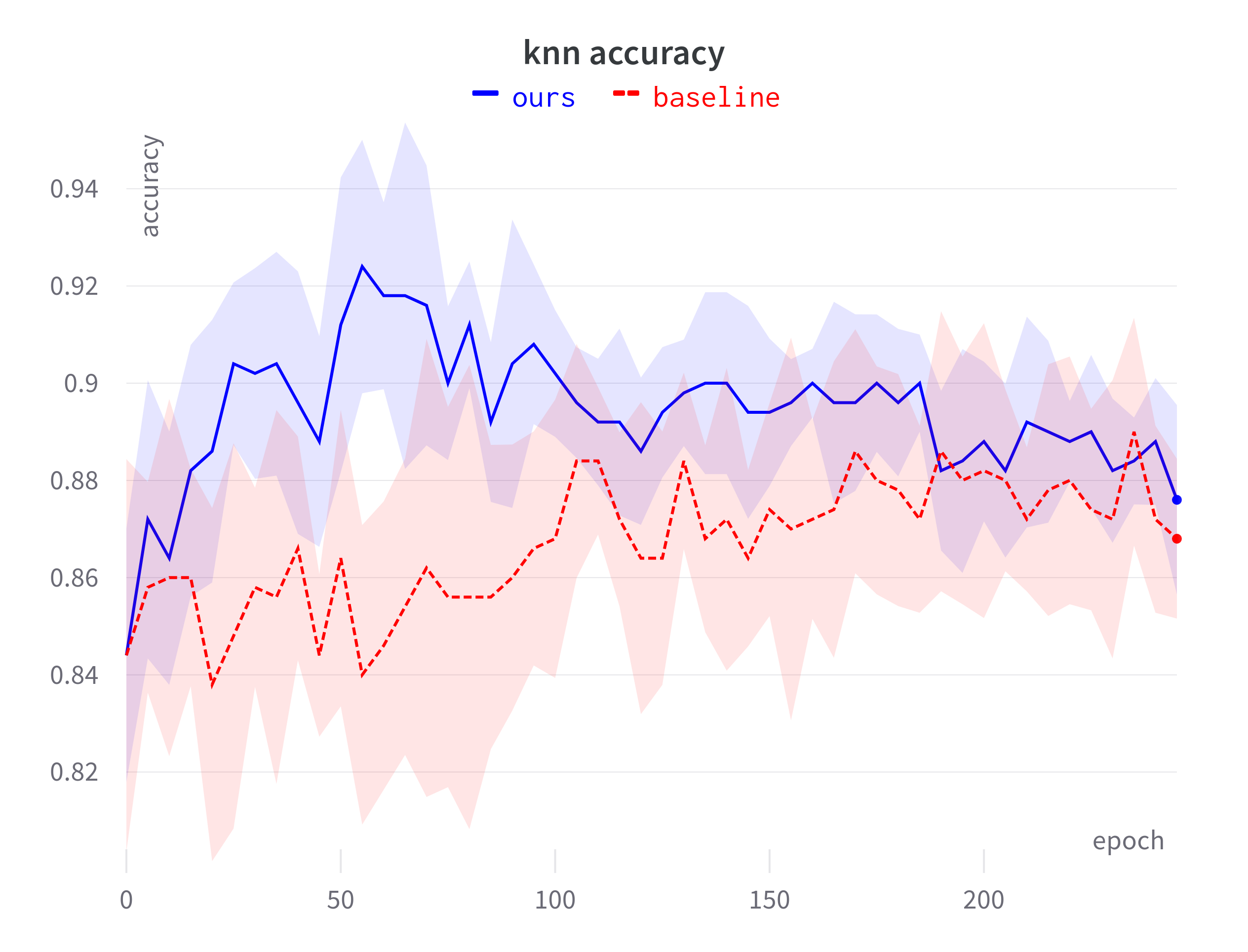}
         \label{fig:acc}
     \end{subfigure}
     \hfill
     \begin{subfigure}[b]{0.3\textwidth}
         \centering
         \includegraphics[width=\textwidth]{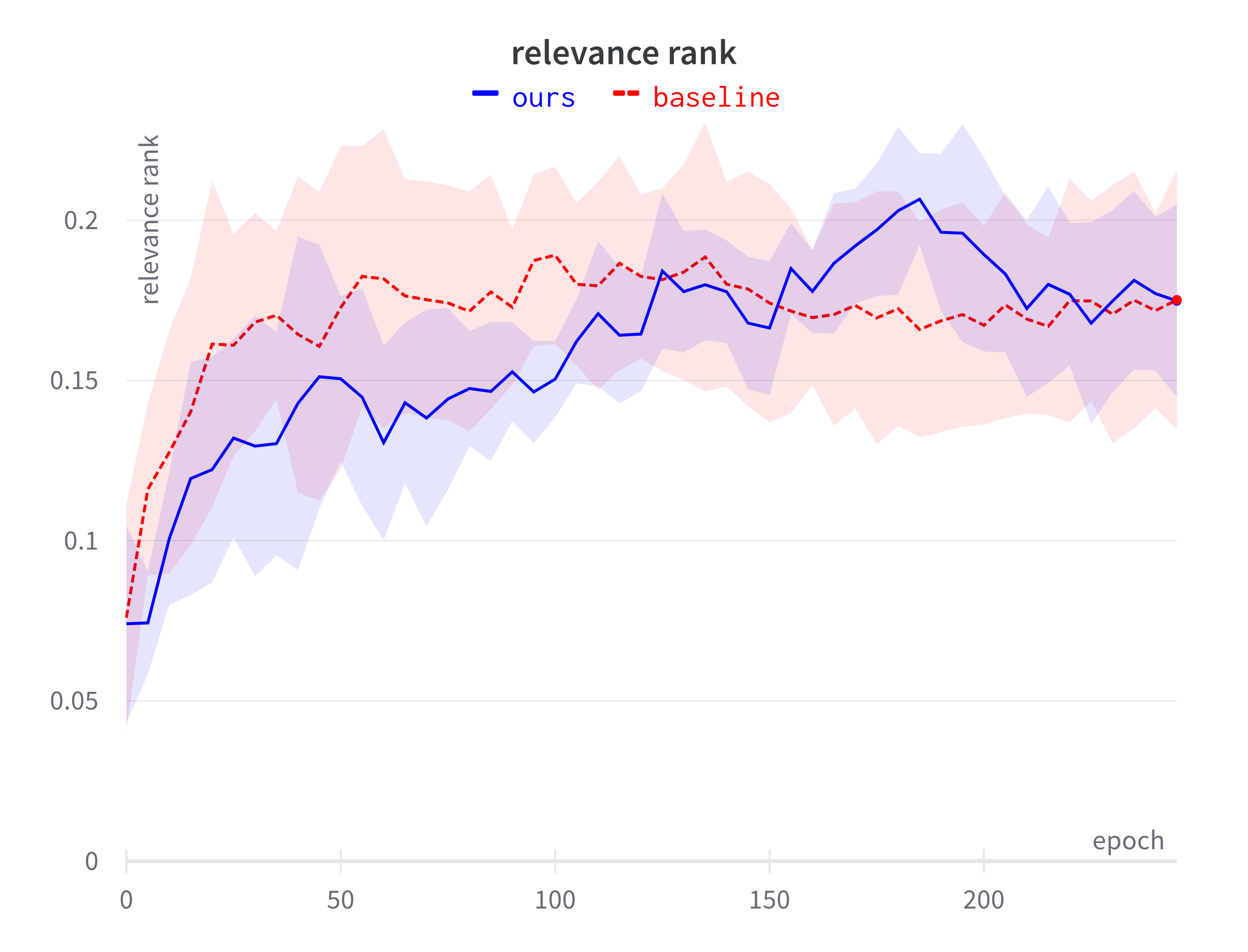}
         \label{fig:rr}
     \end{subfigure}
        \caption{From left to right, mean average precision, knn accuracy, and relevance rank scores versus epochs across 5 training runs on the test images from the UNN dataset. The plot show how performance increase with training time, and that the proposed framework learns faster with better results.}
        \label{fig:unn-scores}
\end{figure*}

\section{Data}

In this section, we present the data used to evaluated our proposed framework.

\subsection{Decathlon data}

The medical segmentation decathlon is a biomedical image analysis challenge where several tasks and modalities are considered \citep{antonelli2021medical}. One of the datasets in this challenge is a CT liver dataset acquired from the IRCAD Hopitaux Universitaires and consists of 201 contrast-ehanced CT liver images from patietns with mostly cancers and metastaic liver disease. However, we exclude 70 of these images as they do not include label information. Using every slide from each volume is computationally intractable. Therefore, we construct a slice-wise dataset as follows. From each volume, we sample 5 slices with no liver and 5 slices with liver. We construct the training set from the first 100 volumes and the test set from the remaining 31 volumes. This results in a balanced dataset with 1310 training images and 310 test images.

\subsection{UNN data}
The UNN dataset is from an extensive database of CT scans from The University Hospital of North Norway (UNN). It is under development through a close collaboration between UiT, The Arctic University of Norway, and UNN. The database contains CT volumes of 376 patients surgically treated for rectal cancer from 2006 to 2011 in North Norway. The examinations were conducted for diagnostic and routine follow-up purposes. The full dataset consists of CT with coronal, sagittal, and axial slices of mainly the thorax, abdomen, and pelvis. Examinations were conducted with different scanners and protocols at eight different hospitals in North Norway in the period 2005 to 2020.

From the full dataset a subset of 3347 axial volumes from 368 patients was selected based on descriptive keywords and DICOM metadata to limit it to contain mostly volumes of the liver and
abdomen. This subset is similar to the CT liver partition of the medical segmentation decathlon dataset in terms of image resolutions and contents, but more diverse in terms of image quality,
contrast enhancement levels, and artifacts because it is only curated using keywords and metadata, and not by manual assessment.

From the UNN subset 10 randomly selected volumes from 10 different and randomly selected patients with liver tumors were manually labeled with segmentation masks of the liver and metastatic regions by a clinician (co-author K.R.) to be used in our study. In addition, two volumes from a patient that had been treated with liver surgery to remove a metastatic liver segment were included. One volume was before the surgery, and one after the surgery. The study of these pre- and post-operative images is conducted as a use-case of cross-examination CBIR.

\section{Experiments}

We present the results of the experimental evaluation of our proposed framework. All models were trained with a batch size of 32 and for 250 epochs. Optimization was carried out using stochastic gradient descent with momentum=0.9, weight decay=0.0001, and learning rate=0.05 * batch size / 256, as used in the SimSiam framework \citep{Simsiam}. As in previous works \citep{simclr, Simsiam}, a ResNet50 \citep{7780459} was used as the feature extractor, with the output of the average pooling layer as the final representation. For both the KNN classifier and the MAP we set K=5. Code is available at \url{https://github.com/Wickstrom/clinical-self-supervised-CBIR-ct-liver.git}.

\subsection{Quantitative results}

Table \ref{tab:dec} and \ref{tab:unn} presents the MAP, accuracy of a 5NN classifier, and the RR on the test data from the Decathlon and UNN datasets. The results show that the proposed framework outperforms the standard self-supervised approach across most scores. Furthermore, self-supervised learning greatly improves upon simply using the feature extractor trained on the Imagenet dataset. Also, the improvements are transferable across datasets, as the feature extractors trained on the Decathlon data also leads to improved performance in the UNN data.

Figure \ref{fig:scores} and \ref{fig:unn-scores} presents the evolution of MAP, accuracy of a 5NN classifier, and the RR on the test data from the Decathlon and UNN datasets across training. The plots highlight how the scores improve as training progresses and stabilizes. However, an interesting observation is that the MAP and KNN accuracy achieves its highest value earlier in training on the UNN dataset. 

\begin{table}[ht]
    \centering
    \caption{Mean and std of mean average precision, knn accuracy and relevance rank score across 5 training runs on the test images from the Decathlon dataset. Results show that the proposed framework outperforms the baselines. Bold numbers indicates the highest performing model.}
    \begin{tabular}{l|ccc}
        \toprule
        {pretraining} & MAP & ACC & RR \\
        \midrule
        IN & 79.4  & 80.3 & 5.00  \\
        IN + SS (baseline) & 87.7 $\pm$ 1.0 & 88.8 $\pm$ 2.0 & 13.6 $\pm$ 2.5 \\
        IN + SS (ours) & \textbf{89.1} $\pm$ \textbf{1.3} & \textbf{90.6} $\pm$ \textbf{1.4} & \textbf{16.2} $\pm$ \textbf{3.1} \\
        \bottomrule
    \end{tabular}
    \label{tab:dec}
\end{table}

\begin{table}[ht]
    \centering
    \caption{Mean and std of mean average precision, knn accuracy and relevance rank score across 5 training runs on test images from the UNN dataset. Results show that the proposed framework outperforms the baselines. Bold numbers indicates the highest performing model.}
    \begin{tabular}{l|ccc}
        \toprule
        {pretraining} & MAP & ACC & RR \\
        \midrule
        IN & 80.7  & 83.0 & 4.34  \\
        IN + SS (baseline) & 84.1 $\pm$ 2.3  & 86.8 $\pm$ 1.6 & \textbf{17.5} $\pm$ \textbf{4.0}\\
        IN + SS (ours) & \textbf{85.1} $\pm$ \textbf{1.7} & \textbf{87.6} $\pm$ \textbf{1.9} & \textbf{17.5} $\pm$ \textbf{3.0} \\
        \bottomrule
    \end{tabular}
    \label{tab:unn}
\end{table}

\begin{figure*}[ht]
    \centering
    \includegraphics[width=0.9\textwidth]{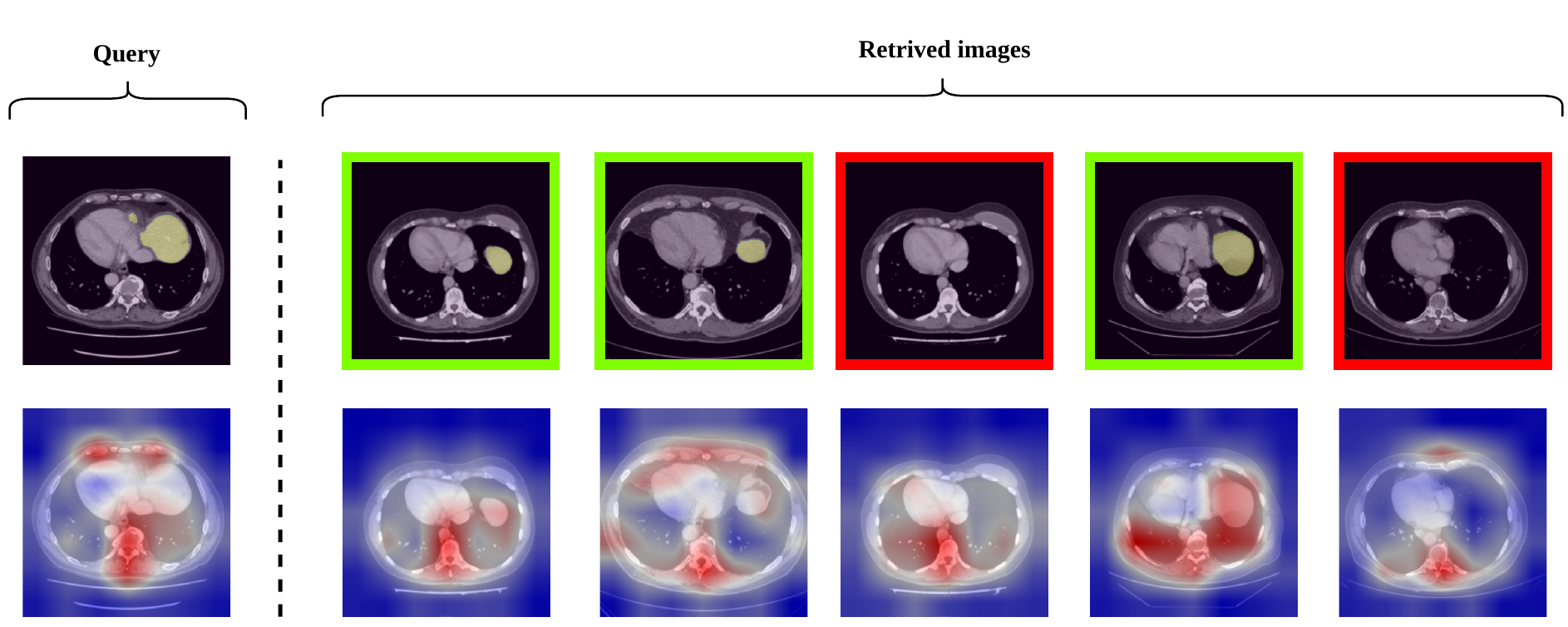}
    \caption{\textbf{Example (1):} CBIR example from Decathlon dataset with feature extractor pretrained on Imagenet dataset. Top row shows, from left to right, the query and the top 5 retrieved images. Bottom row shows the important features for the representation of each image, with important features in red and less important in blue. Some of the retrieved images do not contain the liver, and the explainability analysis shows that the feature extractor is focusing on the spine and rib cage instead of the organs. This information is important to understand why non-relevant images are retrieved, and would not be available without the explainability analysis.}
    \label{fig:in}
\end{figure*}

\subsection{Explaining representations - qualitative results}

The relevance rank scores in Table \ref{tab:dec} and \ref{tab:unn} show that the proposed framework utilizes liver features in the image to a larger degree than the baseline approaches. However, the scores are far from perfect, which means that other parts of the image are also being used. Also, the feature extractor that is only trained on the Imagenet dataset has a very low relevance rank score, meaning that it is putting little attention on the liver. All of these observations can be investigated through XAI. In this section, we illuminate these observations through a new explainability analysis for CBIR by leveraging the RELAX framework that was described in Section \ref{sec:relax}. We show 4 qualitative examples, where the first example shows explanations for the feature extractor trained using Imagenet, and the remaining three examples shows explanations for the feature extractor trained using the proposed framework. In all examples, we show a query from the test set and the 5 retrieved images by CBIR system. Additionally, we show the explanation for the query and retrieved images. The explanation show which features in the input are the most important for the representation of the image, where important pixels are highlighted in red and non-important pixels in blue.

\textbf{Example 1: the feature extractor pretrained on Imagenet focuses on hard edges such as the spine.} Figure \ref{fig:in} displays an example where 2 of the 5 the retrieved images do not contain parts of the liver. When inspecting the explanations, it is clear that the feature extractor is not focusing on the liver, but rather on the tailbone. We hypothesize that since the feature extractor has never been presented with CT images, it utilizes prominent features with hard edges such as spine, as opposed to organs with softer boundaries. The behaviour discovered in this example is important, as it might also result in unexpected or poor retrievals for other queries.

\begin{figure*}[ht]
    \centering
    \includegraphics[width=0.9\textwidth]{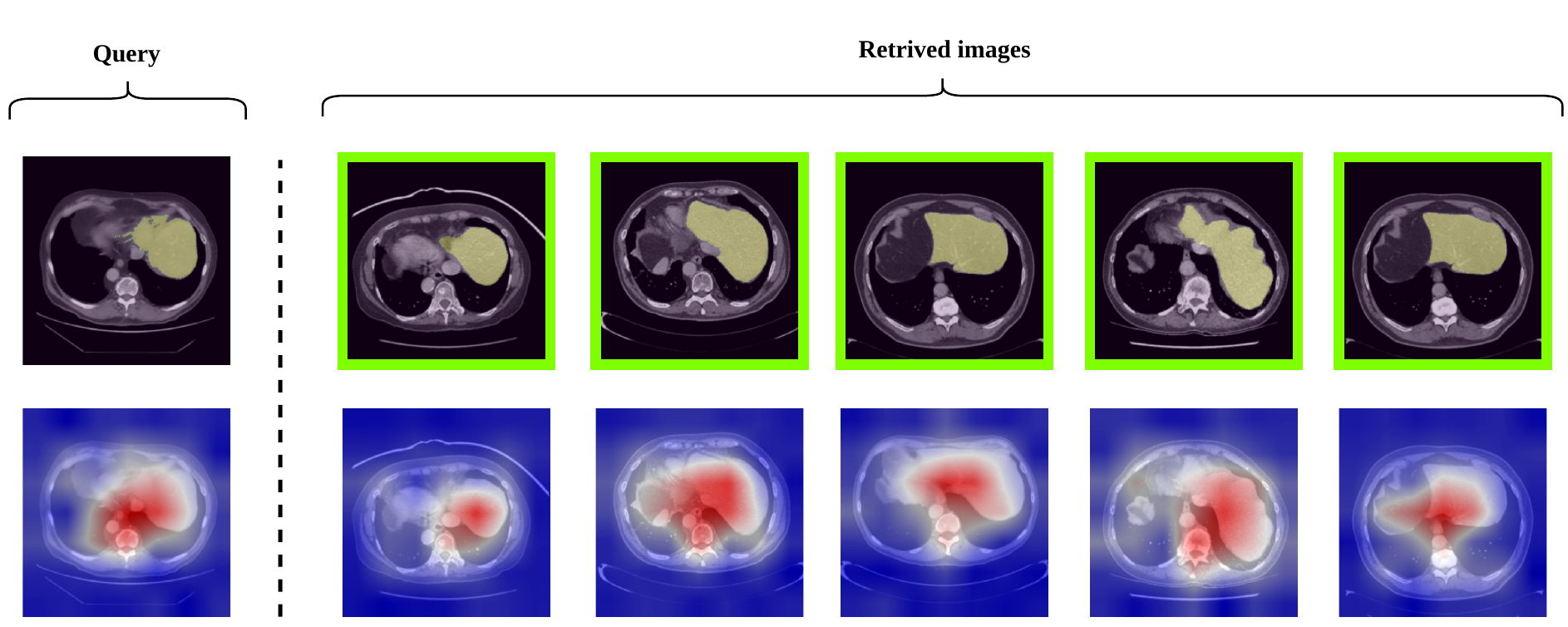}
    \caption{\textbf{Example (2):} CBIR example from Decathlon dataset with feature extractor pretrained using the proposed self-supervised framework. Top row shows, from left to right, the query and the top 5 retrieved images. Bottom row shows the important features for the representation of each image, with important features in red and less important in blue. All retrieved images contain the liver, and the explainability analysis shows that the feature extractor is focusing on the liver.}
    \label{fig:ss}
\end{figure*}

\textbf{Example 2: the feature extractor trained using the proposed framework focuses on liver features.} Figure \ref{fig:ss} shows an example where all the retrieved images contain liver. Additionally, it is evident that the feature extractor is putting more emphasis on the liver for all the images, which illustrates how the proposed self-supervised framework has enabled the feature extractor to focus on clinically relevant features.

\begin{figure*}[ht]
    \centering
    \includegraphics[width=0.9\textwidth]{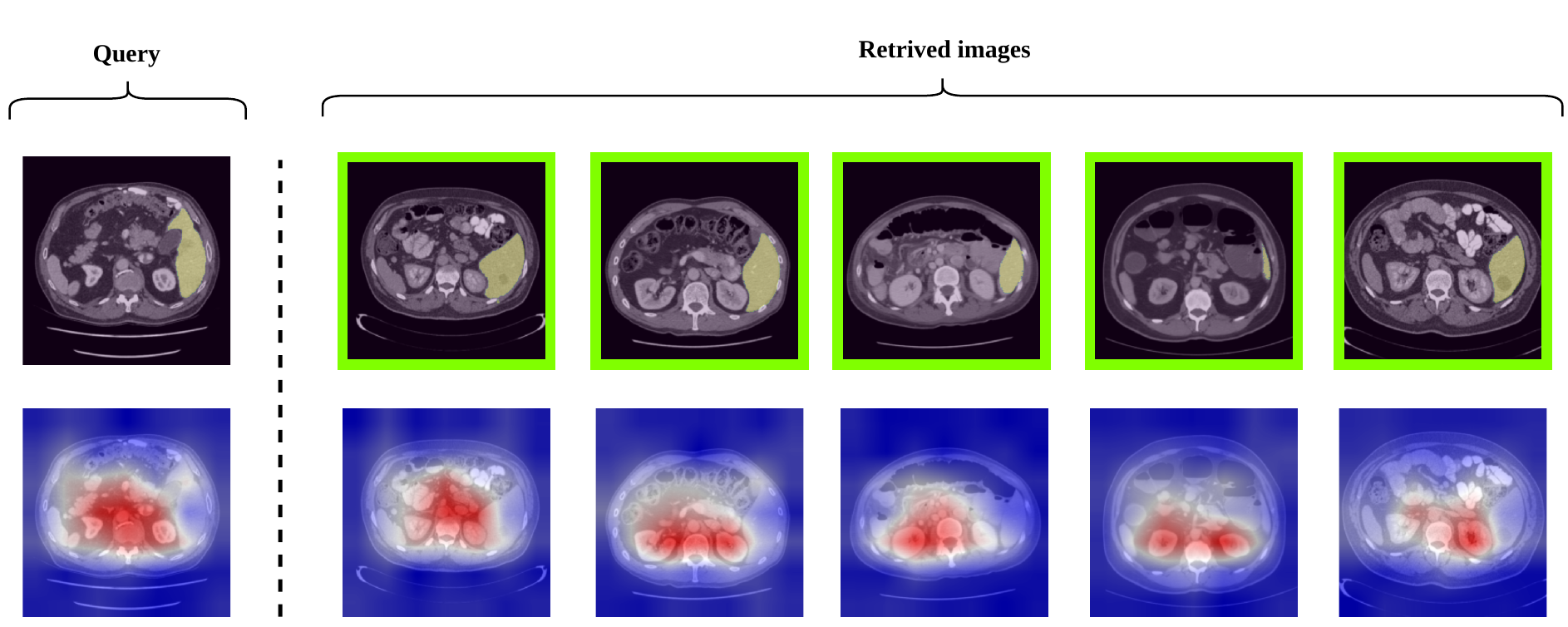}
    \caption{\textbf{Example (3):} CBIR example from Decathlon dataset with feature extractor pretrained using the proposed self-supervised framework. Top row shows, from left to right, the query and the top 5 retrieved images. Bottom row shows the important features for the representation of each image, with important features in red and less important in blue. All retrieved images contain the liver, but the explainability analysis reveals that the focus is on the kidneys, not the liver.}
    \label{fig:ss2}
\end{figure*}

\textbf{Example 3: the feature extractor trained using the proposed framework uses features from organs that often co-occur with the liver.} Figure \ref{fig:ss2} displays an example where CBIR system retrieves 5 images that contain the liver, but where the explainability analysis shows that it not focusing on part of the images where the liver is present. Instead, it puts attention on the kidneys, which are quite prominent in all images. The kidneys often occur together with the liver in many CT images, and it also has similar pixel intensities as the liver (in terms of Houndsfield units). Therefore, it is not surprising that the feature extractor has learned to utilize both liver and kidney features, which also explains the behaviour in this example. Such insights would not be obtainable without conducting the explainability analysis.

\begin{figure*}[ht]
    \centering
    \includegraphics[width=0.9\textwidth]{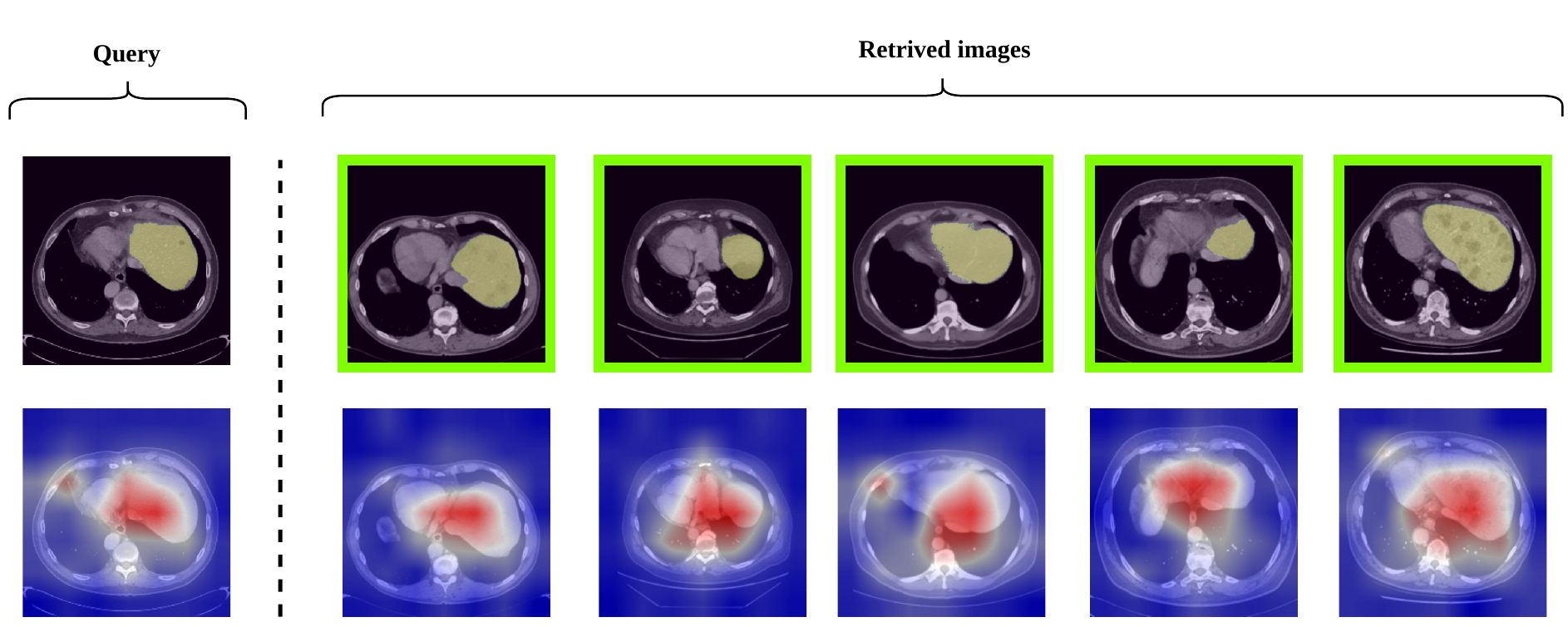}
    \caption{\textbf{Example (4):} CBIR example from UNN dataset with feature extractor pretrained using the proposed self-supervised framework. Top row shows, from left to right, the query and the top 5 retrieved images. Bottom row shows the important features for the representation of each image, with important features in red and less important in blue. All retrieved images contain the liver and the feature extractor is focusing on the liver, which illustrates that the feature extractor trained on the Decathlon dataset transfers well to the UNN dataset.}
    \label{fig:unn}
\end{figure*}

\textbf{Example 4: the feature extractor trained using the proposed framework focuses on liver features, also for images from a different dataset.} Lastly, Figure \ref{fig:unn} shows and example from the UNN dataset. This example illustrates that also on this new and unseen dataset, the feature extractor is basing the representation of these images features associated with the liver. 

\subsection{Case study: cross-examination CBIR}\label{sec:use-case}

A typical scenario in clinical practice is comparing a newly conducted examination with one ore more previous examinations. For instance, one might want to compare a particular slice from the new examination with a selection of slices from one or several previous examinations. Such a comparison can help physicians understand how a diseases has progress since the previous examination, such as the development of liver metastasis. But when conducting such a comparison, the physician must manually inspect the new examination, and potentially several previous examinations. The CT scans are often taken with different settings across examinations, and it is therefore not possible to simply select the same slice from different examinations, as this can image completely different parts of the patient. A precise and reliable CBIR system could make such a cross-examination more efficient, by automating the retrieval process for the physician.

A typical scenario in clinical practice is comparing a newly conducted examination with one or more previous examinations. Such as development, progress, or effect of treatment of liver metastasis, the status of liver cirrhosis, auto-immune diseases in the liver, or any morphological or anatomical changes in the liver over the course. One might want to compare a region of interest in the slice from the current examination with previously conducted examinations. The comparison result will be applicable for evaluating the disease over the long course. In routine clinical practice, the selection of slices from the previous examinations is made manually to achieve the comparison, which is very time-consuming. CT scans taken over time are often conducted in an inconsistent sequence and contain unlike body regions in the same array of slices; therefore, selecting the identical arrays of slices from the different examinations is inaccurate. A precise and transparent CBIR system could make a cross-examination more efficient through the automatic retrieval process.

\begin{figure*}
    \centering
    \includegraphics[width=0.9\textwidth]{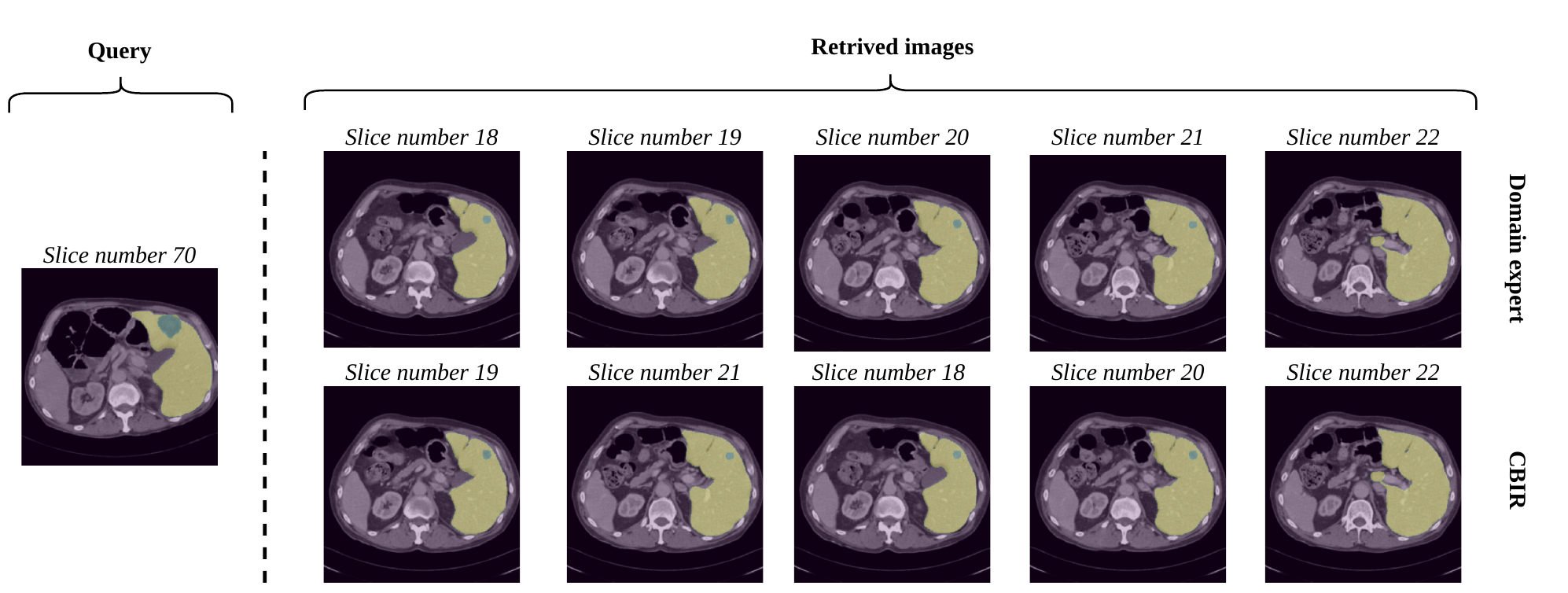}
    \caption{An example of cross-examination CBIR. The query is from a recent examination, and the retrieved images are from a prior examination from the same patient. The goal of such a study is to investigate the development of liver metastasis. The query and retrieved images in the top row are selected by an experienced physician, and the bottom row are the retrieved images from the CBIR system. The CBIR successfully retrieves the same images as the physician, but lacks the spatial coherence to order the retrieved images.}
    \label{fig:case-study}
\end{figure*}

Figure \ref{fig:case-study} displays an example of such a cross-examination. The query is selected from a recent examination, and the retrieved images are from the previous examination of the same patient. This patient is from the UNN dataset and was selected since liver metastasis has been developed between the two examinations. The query was selected by an experienced physician (co-author K.R.), which also selected five images to examine from the previous examination. Ideally, the CBIR system should align well with the image selected by the physician. In this example, the CBIR system produces a successful retrieval, as it identifies the same images as the physician.
However, an interesting observation is that the CBIR retrieved are not sorted in the same manner as the physician’s retrievals. Probably this deviation is due to the CBIR system being trained on single slices without a sense of spatial coherence. Future works could address this by incorporating neighboring samples as positive pairs in the self-supervised training.

\section{Conclusion}
We propose a clinically motivated self-supervised framework for CBIR of CT liver images. Our proposed framework exploits the properties of the liver to learn more clinically relevant features, which results in show leads to improved performance. Moreover, we leverage the RELAX framework to provide the first representation learning explainability analysis in the context of CBIR of CT liver images. Our analysis provides new insights into the feature extraction process and shows how self-supervised learning can provide feature extractors that extract more clinically relevant features compared to feature extractors trained on non-CT liver images. Our experimental evaluation also shows how the proposed framework generalizes to new datasets, and we present a clinically relevant user study. In future works, we intend to investigate how the proposed approach can be extended to extract features specific to other organs based on clipping strategies catered specifically to the desired organ. We believe that the proposed framework can play an essential role in constructing reliable CBIR that can effectively utilize unlabeled data.

\section*{Declaration of competing interest}
The authors declare that they have no known competing financial interests or personal relationships that could have appeared to
influence the work reported in this paper.

\section*{Acknowledgments}
This work was supported by The Research Council of Norway
(RCN), through its Centre for Research-based Innovation funding
scheme [grant number 309439] and Consortium Partners; RCN
FRIPRO [grant number 315029]; RCN IKTPLUSS [grant number
303514]; and the UiT Thematic Initiative.

%%Harvard
\bibliographystyle{model2-names.bst}\biboptions{authoryear}
\bibliography{refs}

\end{document}